%% file: ms.tex
\documentclass{article}

\usepackage{arxiv}
\pdfoutput=1 

\usepackage{url}

\usepackage[T1]{fontenc}
\usepackage{lmodern}
\usepackage{hyperref}       
\usepackage{url}            
\usepackage{booktabs}       
\usepackage{amsfonts}       
\usepackage{nicefrac}       
\usepackage{microtype}      
\usepackage{lipsum}		
\usepackage{hyperref}
\usepackage{array}
\usepackage{subfigure}
\usepackage{amsmath}
\usepackage{amssymb}
\usepackage{bbm}
\usepackage{tikz}
\usepackage{siunitx}
\usepackage{algorithm}
\usepackage{algpseudocode}
\usepackage{todonotes}
\usepackage{tikz}
\usetikzlibrary{fit,positioning}

\usepackage{amsmath}
\DeclareMathOperator*{\argmax}{arg\,max}

\title{\LARGE \bf
Rapidly adapting robot swarms with Swarm Map-based Bayesian Optimisation
}
\author{David M. Bossens and Danesh Tarapore$^{}$
\thanks{$^{}$ Authors are with School of Electronics and Computer Science, University of Southampton, SO17 1BJ Southampton, U.K. {\tt\small d.m.bossens@soton.ac.uk}}%
}

\begin{document}

\maketitle
\thispagestyle{empty}
\pagestyle{empty}

\begin{abstract}
Rapid performance recovery from unforeseen environmental perturbations remains a grand challenge in swarm robotics. To solve this challenge, we investigate a behaviour adaptation approach, where one searches an archive of controllers for potential recovery solutions. To apply behaviour adaptation in swarm robotic systems, we propose two algorithms: (i) Swarm Map-based Optimisation (SMBO), which selects and evaluates one controller at a time, for a homogeneous swarm, in a centralised fashion; and (ii) Swarm Map-based Optimisation Decentralised (SMBO-Dec), which performs an asynchronous batch-based Bayesian optimisation to simultaneously explore different controllers for groups of robots in the swarm. We set up foraging experiments with a variety of disturbances: injected faults to proximity sensors, ground sensors, and the actuators of individual robots, with 100 unique combinations for each type. We also investigate disturbances in the operating environment of the swarm,  where the swarm has to adapt to drastic changes in the number of resources available in the environment, and to one of the robots behaving disruptively towards the rest of the swarm, with 30 unique conditions for each such perturbation. The viability of SMBO and SMBO-Dec is demonstrated, comparing favourably to variants of random search and gradient descent, and various ablations, and improving performance up to 80\% compared to the performance at the time of fault injection within at most 30 evaluations.
\end{abstract}

\input{introduction.tex}

\input{methods.tex}

\input{experimental_setup.tex}

\input{results.tex}

\input{conclusion.tex}

\bibliographystyle{IEEEtran} 
\bibliography{biblio} 

\end{document}

%% file: introduction.tex
\section{Introduction}
Robotic systems have the potential to solve tasks that are too dangerous, repetitive, or laborious for humans may be done instead by robots. Multi-robot and swarm robotic systems are often preferred over single-robot systems due to providing modular solutions to problems, for example, they can gather information in a distributed manner, cover large areas, and perform cooperative tasks \cite{Sahin2005}. Another major advantage of robot swarms is their redundancy, allowing the swarm to continue functioning despite failures in one or more robots of the swarm  \cite{Brambilla2013a}. Despite this redundancy, it is preferable to allow robots to adapt their behaviour to recover from sustained damages, as robots are expensive, may carry valuable data about their mission, and may pose environmental concerns if left unrecovered post-mission.\\
\indent Traditionally, there are two approaches for fault tolerance in robot swarm. Data-driven abnormality detection methods  identify robots that behave abnormally due to sustained faults \cite{Chandola2009}. While this approach can successfully identify faulty robots in a swarm \cite{Christensen2009,Tarapore2015,Tarapore2019} it does not provide diagnostic information, and therefore provides no straight-forward solution to fault \textit{recovery}. By contrast, in model-based fault detection, one typically makes a number of models of faulty behaviours -- one for each fault --  and tracks how closely these behaviours match with reality. Model-based fault detection systems have been deployed successfully to detect pre-specified faults to the sensors and actuators of mobile robots \cite{Skoundrianos2004,Christensen2008a}. However, the faults have to be anticipated by the designer \textit{a priori}. Moreover, this approach requires an accurate model of the robot, which may not be available for robots operating for extended periods of time. \\
\indent Fault recovery may instead be accomplished via rapid behavioural adaptation, a trial-and-error learning variant in which a large behavioural repertoire evolved by a quality-diversity algorithm  (e.g., \cite{LehmanStanley2011,Mouret2015}) is searched to find the best compensatory behaviour, which makes it possible for the swarm to adapt rapidly without any knowledge of the fault. Behaviour adaptation has been explored successfully for recovering from actuator damages -- for example, in the gait of a hexapod robot and in a pick-and-place task for a multi-joint planar robot arm (for various robot models see \cite{Cully2015b,Bossens2020a}) -- but remains to be investigated in swarm robotics.\\
\indent Our previous work on employing a repertoire of swarm behaviors for fault recovery has demonstrated that quality-diversity algorithms can reduce the impact of the fault by a factor 2-3 by performing exhaustive search across the evolved behavioural repertoire \cite{Bossens2020}. However, this study had three limitations: (i) the exhaustive search is too expensive since in practice only a limited number of performance evaluations can be granted; (ii) the use of a homogeneous swarm limits the adaptivity when different robots in the swarm are affected by different faults; and (iii) in addition to faults to sensors or actuators, other types of faults or even environmental perturbations may potentially disrupt the behaviour of the robot swarm. In this paper, we conduct a proof-of-concept study for rapid behaviour adaptation in robot swarms, investigating how to overcome these limitations and efficiently recover from a variety of unforeseen perturbations. Due to its relevance for transport and, more generally, loosely-coordinated swarms, we investigate a foraging task, a commonly used benchmark for swarm robotic systems in which robots must search and grab food and then bring them back to an area called the nest \cite{Bayindir2016}.

%% file: methods.tex
\section{Rapid adaptation with Swarm Map-Based Bayesian Optimisation}
To allow a swarm of robotic agents to adapt to unforeseen perturbations in the environment or damages to any of the robots' sensors or actuators, we make use of a two-phase approach that extends the Intelligent Trial and Error (IT\&E) algorithm \cite{Cully2015b} by accounting for communication and behavioural requirements in swarms and exploiting decentralised Bayesian optimisation for rapid behavioural adaptation across the swarm. First, in simulation, MAP-Elites \cite{Mouret2015} evolves a behaviour-performance map, a large archive of behaviourally diverse high-performing solutions. Second, after an observed drop in task performance in the target application of the robot swarm, a Bayesian optimisation algorithm, called Swarm Map-based Bayesian Optimisation (SMBO), searches rapidly for the best compensatory behaviour in the behaviour-performance map, using the performances in the behaviour-performance map as Bayesian priors.

\subsection{Evolving Bayesian priors with MAP-Elites}
To capture the variety of swarm behaviours that can be used for adaptation and their corresponding prior performance estimate, we use MAP-Elites \cite{Mouret2015} to evolve a behaviour-performance map, an archive storing for each local region in a behaviour space the best solution and its associated performance.

The MAP-Elites algorithm first creates an initial population of random genotypes. Each such genotype $\mathbf{g}$ represents the controller $C(\mathbf{g})$  of all of the robots in a homogeneous swarm. The homogeneous swarm is evaluated on several evaluation trials that result in a fitness score $f$ and a behavioural descriptor $\mathbf{x} \in [0,1]^D$, a low-dimensional description of the behaviour of the robot collective in the environment. The behaviour-performance map is then populated according to the following replacement rule: if the bin corresponding to the individual's behavioural descriptor is empty (i.e., $\mathcal{M}[\mathbf{x}]=\emptyset$) or if the new individual has a higher fitness than the solution in that bin (i.e., $ f(\mathbf{g}) > f(\mathcal{M}[\mathbf{x}])$), place the individual's genotype $\mathbf{g}$ and its performance in that bin of the behaviour-performance map (i.e., $\mathcal{M}[\mathbf{x}] \gets \mathbf{g}$).

After initialisation, the algorithm repeats iterations that include random selection, genetic variation, evaluation and replacement. Random selection is implemented by selecting genotypes at random from non-empty bins in the behaviour-performance maps. Genetic variation is applied using a mutation operator. Evaluation and replacement follow the same rules as in the above-mentioned initialisation. After many repetitions of this cycle, the behaviour-performance map is gradually filled with a behaviourally diverse map of high-performing solutions.

\subsection{Swarm Map-based Bayesian Optimisation (SMBO)}
\label{sec: SMBO}
Swarm-MBO (SMBO) employs a centralised Gaussian process model shared across the swarm from which controllers are selected. To recover performance from faults or other adverse environmental perturbations, we model the performance $f$ as a function of the behavioural descriptor $\mathbf{x}$ to search for the best solution across the evolved behaviour-performance map. SMBO selects a single controller, communicates it to the swarm such that each of its robots will then use a copy of this controller, evaluates the fitness of the swarm in the environment, and then updates the shared Gaussian process model.

SMBO starts by initialising the prior based on the behaviour-performance map $\mathcal{M}$. Each behavioural bin in the map constitutes a data point $\mathbf{x}$ such that
\begin{equation}
\label{eq: prior}
f(\mathbf{x}) \sim \mathcal{N}\!(\mu_0(\mathbf{x}), \sigma_0^2(\mathbf{x}))  \quad \forall \mathbf{x} \in \mathcal{M} \,,
\end{equation}
where $\mu_0(\mathbf{x}) = f(\mathcal{M}[\mathbf{x}])$, the performance obtained in evolution, and $\sigma_0^2(\mathbf{x}) = k(\mathbf{x},\mathbf{x}) + \sigma_{noise}^2$, the variance according to the kernel function and a user-defined noise, serve as the mean and variance, respectively, of the Gaussian distribution. We use the Matern kernel (with $\nu=5/2$):
\begin{equation}
\begin{aligned}
k(\mathbf{x}_i,\mathbf{x}_j) = \left( 1+ \frac{\sqrt{5}}{\rho} ||\mathbf{x}_i - \mathbf{x}_j||_2
+ \frac{5}{3\rho^2} ||\mathbf{x}_i - \mathbf{x}_j||_2^2 \right) \\ \exp\left(- \frac{\sqrt{5}}{\rho}||\mathbf{x}_i - \mathbf{x}_j||_2\right) \,,
\end{aligned}
\end{equation}
where $\rho$ is a user-defined parameter, called the length-scale.

After initialisation, a cycle of multiple iterations $t=1,\dots,T$ is repeated, each of which consist of four steps.

The first step is to sample a new point $\mathbf{x}_t \in \mathbb{R}^D$ by exhaustively searching for the best, unexplored controller -- that is, the behaviour $\mathbf{x}_t \in \mathcal{M} \setminus \{\mathbf{x}_1,\dots,\mathbf{x}_{t-1}\}$ for which $f(\mathbf{x}_t)$ is maximal. The estimate of the best controller is based on the Upper Confidence Bound (UCB) acquisition function. UCB approaches the exploration-exploitation dilemma using an optimistic estimate of performance that accounts for uncertainty in the Gaussian process: 
\begin{equation}
\label{eq: UCB}
\mathbf{x}_t  = \argmax_{\mathbf{x} \in \mathcal{M} \setminus \{\mathbf{x}_1,\dots,\mathbf{x}_{t-1}\}} \mu_{t-1}(\mathbf{x}) + \alpha \sigma_{t-1}(\mathbf{x})\,,
\end{equation}
where $\mu_{t-1}(\mathbf{x})$ and $\sigma_{t-1}(\mathbf{x})$ are the mean and standard-deviation of the Gaussian process, respectively, for a given point $\mathbf{x}$, at evaluation $t-1$; and $\alpha > 0$ represents the exploration-exploitation dilemma.

The second step is to evaluate the performance of the newly sampled behaviour. The controller corresponding to the sampled point, $c = C(\mathcal{M}[\mathbf{x}_t])$, is obtained from the behaviour performance map, and is then communicated to all the robots in the swarm. As a homogenous swarm, the robots then perform their task, each with their own copy of controller $c$, after which their collective performance $f(\textbf{x}_t)$ is assessed.

As a third step, the Gaussian process is updated with the newly sampled point $\mathbf{x}_t$. The SMBO algorithm updates the model based on the difference between the behaviour-performance map obtained from simulation and the performance in the real world. For behaviours  $\mathbf{x} \in \mathcal{M}$, the mean is updated according to 
\begin{equation}
\mu_t(\mathbf{x}) = f(\mathbf{x})+
\mathbf{k}^{\intercal} \mathbf{K}^{-1} 
\left(f_{1:t} -
P_{1:t}) \right) \,,
\end{equation}
where $\mathbf{x}_{1:t}$ are the sampled points, $f_{1:t}$ their corresponding performances, and $P_{1:t}$ their corresponding priors (the 
performances obtained in the unperturbed environment); $\mathbf{K} \in \mathbb{R}^{t \times t}$ computes for each pair of observations their covariance 
using the Matern kernel function $k$ while accounting for noise in observations by adding a diagonal matrix with a fixed noise variance, $\sigma_{noise}^2 \mathbf{I}$; and $\mathbf{k} \in \mathbb{R}^{t \times 1}$ similarly computes the covariance between the queried point $\mathbf{x}$ and the samples in $\mathbf{x}_{1:t}$.

Finally, the variance of the Gaussian process is updated:
\begin{equation}
\sigma_t^2(\mathbf{x}) = k(\mathbf{x},\mathbf{x}) - 
\mathbf{k}^{\intercal} \mathbf{K}^{-1} \mathbf{k}\,.
\end{equation}

\subsection{SMBO-Decentralised}
\label{sec: SMBO-Dec}
We further propose a second MBO-based algorithm called SMBO-Decentralised for faulty recovery in swarms. It differs in two critical aspects from SMBO. First, SMBO evaluates a single controller using a centralised GP model shared across the swarm; by contrast, in SMBO-Decentralised, each robot in the swarm can serve as a separate worker for a common Gaussian process model. Second, SMBO-Decentralised shares a separate Gaussian process model within groups of robots with the same fault condition, making it possible to compose a heterogeneous swarm that is adapted to the different faults sustained by its robots.

After initialisation of the prior, the algorithm starts adding new samples $\mathbf{x}$ and observations $f(\mathbf{x})$ using batch Bayesian optimisation with local penalisation \cite{Gonzalez2016}, allowing additionally for asynchronicity\footnote{To allow reliable evaluation and simplified analysis, experimental results will assume all robots in the swarm reset synchronously after finishing each trial.}. 

All robots in the swarm initially have the same Gaussian process model, initialised with the evolved behaviour-performance map as in Eq. \ref{eq: prior}, but act as different workers each evaluating a different sample (i.e., a different controller). Each observation represents the \textit{current} estimate of the performance of a controller for a robot of the swarm. Whenever a robot finishes a trial within a set of performance trials of a controller,  the robot updates its Gaussian process model with this sample and its estimated performance, and then broadcasts these to other robots of the swarm that have the same fault condition. To incorporate a greater uncertainty for evaluations if they have a lower number of trials or a larger variance between trials, the observation noise depends on the standard-error $\sigma_{n}^2$ of the queried sample:
\begin{equation}
\mathbf{K} = 
\begin{pmatrix}
k(\mathbf{x}_1,\mathbf{x}_1) + \sigma_{n}^2(\mathbf{x}_1)  & \cdots & k(\mathbf{x}_1,\mathbf{x}_t) \\
k(\mathbf{x}_2,\mathbf{x}_1)  & \cdots & k(\mathbf{x}_2,\mathbf{x}_t) \\
\vdots   & \ddots & \vdots  \\
k(\mathbf{x}_t,\mathbf{x}_1) & \cdots & k(\mathbf{x}_t,\mathbf{x}_t) + \sigma_{n}^2(\mathbf{x}_t) \\
\end{pmatrix} \,,
\end{equation}
where $t$ is the number of observations that have one or more trials completed; and the standard error is computed as the variance across trials divided by the number of trials -- if only one trial has been completed then  $\sigma_{n}^2 \gets f(\mathbf{\tilde{x}})^2$.

To ensure that samples in a batch are informative and sample distinct regions of the behaviour space, local penalisation adds a penalty to UCB:
\begin{equation}
\label{eq: UCB-local penalisation}
\mathbf{x}_t  = \argmax_{\mathbf{x} \in \mathcal{M} \setminus \{\mathbf{x}_1,\dots,\mathbf{x}_{t-1}\}} \left[\mu_{t-1}(\mathbf{x}) + \alpha \sigma_{t-1}(\mathbf{x}) \right] \prod_{i=1}^{N_b} \phi(\mathbf{x}, \mathbf{\tilde{x}}_i) \,,
\end{equation}
where $\phi(\mathbf{x}, \mathbf{\tilde{x}}_i)$ provides a penalty if $\mathbf{x}$ is close to $\mathbf{\tilde{x}}_i$  and $N_b$ is the number of busy samples, those currently being evaluated by workers. More precisely, the penalty function is defined as:
\begin{equation}
\phi(\mathbf{x}, \mathbf{\tilde{x}}_i)= \text{erfc}\big(-\frac{1}{\sqrt{2}\sigma_{t-1}(\mathbf{x})} L || \mathbf{x} - \mathbf{\tilde{x}}_i || - M + \mu_{t-1}(\mathbf{x}) \big) \,,
\end{equation}
where $\text{erfc}$ is the complementary error function, $M$ is the estimated maximum of $f$, and $L=\max_{\mathbf{x} \in \mathcal{M}}|| \nabla f(\mathbf{x})||$ is a valid Lipschitz constant for the behaviour space such that $|| f(\mathbf{x}') - f(\mathbf{x}'') || \leq L || \mathbf{x}' - \mathbf{x}'' ||$ for all $\mathbf{x}',\mathbf{x}'' \in \mathcal{M}$. The penalty function $\phi(\mathbf{x}, \mathbf{\tilde{x}}_i)$ represents the probability that the proposed sample $\mathbf{x}$ is \textit{not} within a ball of radius of $r_i = \frac{M - f(\mathbf{\tilde{x}}_i)}{L}$ from the busy sample $\mathbf{\tilde{x}}_i$, with penalty values progressively closer to $0$ as $\mathbf{x}$ gets closer to $\mathbf{\tilde{x}}_i$.

%% file: experimental_setup.tex
\section{Experimental Setup}

\subsection{Robot swarm and its foraging task}
We use a physics-based, discrete-time robot swarm simulator called ARGoS \cite{Pinciroli2012}, designed to realistically simulate robot swarms. Experiments employ the Thymio robot (\cite{Mondada:223049}), a differential-drive mobile robot that has a length of \SI{11}{cm}, a width of \SI{8.5}{cm}, a maximum linear speed of \SI{10}{cm/s}, a maximum angular speed of $\SI{127.32}{^\circ/s}$, and a control cycle of \SI{0.20}{s}. The Thymio robot model is equipped with five frontal and two rear infrared proximity sensors of range \SI{11}{cm} for obstacle avoidance, two ground sensors that detect grey scale on the floor, and two actuators that control the left and right wheel.

\begin{figure}
\centering
\begin{tikzpicture}[scale=0.08,rotate=90]
  \draw [fill=black!30] (0,0) rectangle ++(21cm,42cm);
  \draw[fill=white] (0,0) rectangle ++(3.2cm,42cm);
  \draw[dashed,->] (-5cm,22cm) -- (1.6cm,22cm);
  \node[below] at (-5cm,22cm) {\footnotesize nest};
  \draw[fill=black!255] (8cm,12cm) circle (1cm);
  \draw[fill=black!255] (8cm,5cm) circle (1cm);
  \draw[fill=black!255] (13cm,10cm) circle (2cm);
  \draw[fill=black!255] (15cm,5cm) circle (2cm);
  \draw[fill=black!255] (16cm,17cm) circle (3cm);
   \draw[fill=black!255] (8cm,33cm) circle (1cm);
  \draw[fill=black!255] (8cm,26cm) circle (1cm);
  \draw[fill=black!255] (13cm,31cm) circle (2cm);
  \draw[fill=black!255] (15cm,26cm) circle (2cm);
  \draw[fill=black!255] (16cm,38cm) circle (3cm);
  \draw[dashed,->] (25cm,19cm) -- (19.2cm,17cm);
  \node[above] at (25cm,19cm) {\footnotesize food source};
  \node[rotate=45] at (15cm,30cm) {\includegraphics[width=0.165cm,height=.165cm]{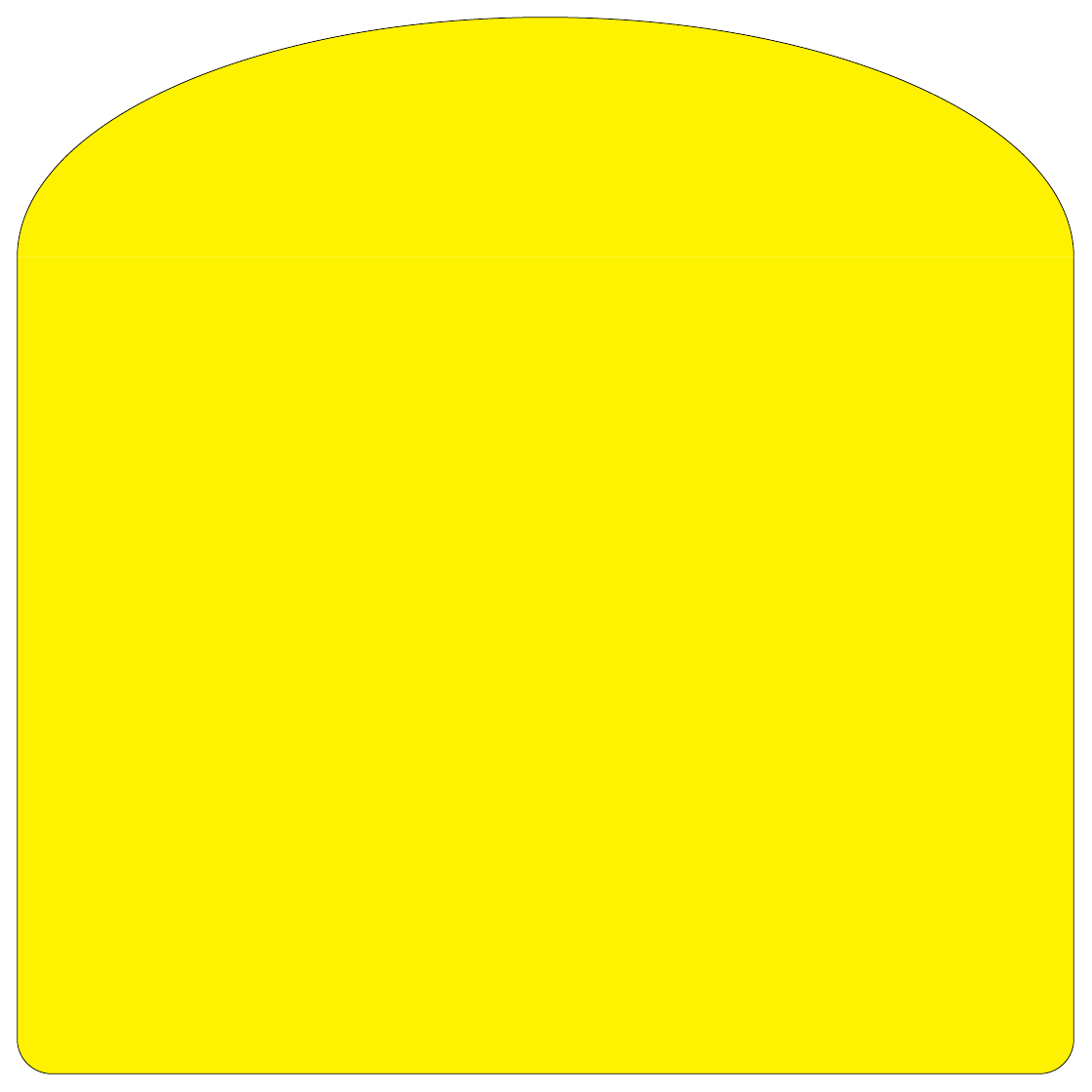}};
  \node[rotate=60] at (10cm,29cm) {\includegraphics[width=0.165cm,height=.165cm]{figures/thymio_case.pdf}};
  \node[rotate=30] at (11cm,16cm) {\includegraphics[width=0.165cm,height=.165cm]{figures/thymio_case.pdf}};
  \node[rotate=10] at (12cm,18cm) {\includegraphics[width=0.165cm,height=.165cm]{figures/thymio_case.pdf}};
  \draw[dashed,<-] (5.0cm,7.0cm) -- (-5cm,-1.5cm);
  \node[below right] at (-5cm,-1.5cm) {\footnotesize robot};
  \node[rotate=10] at (6cm,8cm) {\includegraphics[width=0.165cm,height=.165cm]{figures/thymio_case.pdf}};
  \node[rotate=10] at (6cm,15cm) {\includegraphics[width=0.165cm,height=.165cm]{figures/thymio_case.pdf}};
  \node[above,rotate=90] at (17.5cm,43.4cm) {\small $\SI{2.1}{m}$};
  \draw[<->] (0,43cm) -- (21cm,43cm)  ;
  \node[above] at (22.4cm,4.5cm) {\footnotesize $\SI{4.2}{m}$};
  \draw[<->] (22,0cm) -- (22cm,42cm)  ;
  \node[above,rotate=270] at (1.5cm,-1.4cm) {\footnotesize $\SI{0.32}{m}$};
  \draw[<->] (0cm,-1cm) -- (3.2cm,-1cm)  ; 
\end{tikzpicture}
\caption{\small{Illustration of the foraging task, where 6 robots are initially positioned randomly in the arena and then must repeatedly pick up a food source and bring it to the nest. Food sources have a radius of $\SI{10}{cm}$, $\SI{20}{cm}$, or $\SI{30}{cm}$ and are fixed to the illustrated positions.}} \label{fig: foraging}
\end{figure}

We choose a foraging as the task for the robot swarm, due to its prominence as a benchmark in evolutionary robotics and due to the variety of relevant applications (e.g., pickup-and-delivery of resources of interest). Six Thymio robots are positioned randomly in a $4.2 \times 2.1 ~\SI{}{m^2}$ arena and are tasked to find food items and deliver them to the nest (see Fig. \ref{fig: foraging}).

Each robot in the swarm is controlled by a neural network controller. The controller takes as inputs the 7 proximity sensor readings, the 2 ground sensor readings, and a bias activation, and outputs the left and right wheel velocities of the robot. Robots' ground sensors detect food sources by their black color and the nest by its white color, while the rest of the environment is detected as grey. All sensory activations input to the neural network are scaled in the range $[-1,1]$, while the output neuron activations in range $[-1,1]$ are mapped into wheel velocities in the range $[-10,10] ~\SI{}{cm/s}$.

\subsection{Evolution experiments}
During a first, evolutionary phase of the experiments, MAP-Elites evolves a behaviour-performance map, consisting of a wide diversity of controllers, by applying mutations (weight changes or topology changes) to a directly encoded recurrent neural network. Two different behavioural descriptors are included in the study: (i) the Hand-coded Behavioural Descriptor (\textbf{HBD}), a hand-coded descriptor that tracks statistics of the arena as  a grid of cells \cite{Bossens2020}; and (ii) Systematically Derived Behavioural Characterisations (\textbf{SDBC}) \cite{Gomes2014}, records the mean and standard deviation (across control cycles) of statistics on the relation between the different entities in the task (robots and arena walls). We allow 20,000 generations of evolution with 500 controllers being evaluated per generation\footnote{A single replicate required about \SI{24}{h} of computational time on a 40-cores Intel Xeon Gold 6138 at 2\SI{}{GHz}}. A detailed summary of the parameters for evolution and the behavioural descriptors is given in Section S1 of the Supplementary Information\footnote{Supplementary Information is available from \url{http://tiny.cc/SwarmMBO}.}.

\subsection{Adaptation experiments}
\label{sec: perturbationrecoverysetup}
In the second phase, the following disturbances are injected:
\begin{itemize}
\item \textbf{Proximity-sensor}: each robot is randomly assigned one of four faults to the front proximity sensors, each with probability 0.25: (i) a random value is generated anew at each control cycle; (ii) the sensor value is always maximal; (iii) the sensor value is always minimal; and (iv) there is no fault at all.
\item \textbf{Ground-sensor}: each robot is randomly assigned one of four ground sensor faults, each with probability 0.25: (i) a random value is generated anew at each control cycle; (ii) the sensor value is always maximal; (iii) the sensor value is always minimal value; and (iv) there is no fault at all.
\item \textbf{Actuator}: each robot is randomly assigned one of four actuator faults, each with probability 0.25: (i) the left wheel's speed is halved; (ii) the right wheel's speed is halved; (iii) both wheel's speeds are halved; and (iv) there is no fault at all.
\item \textbf{Software-Nest}: one robot is put on the border of the nest and moves back and forth randomly, with complete loss of its normal control algorithm; 
\item \textbf{Software-Food}: one robot is put on a randomly chosen food source and cannot move, with complete loss of its normal control algorithm; 
\item \textbf{Food-scarcity}: only a single food source remains in the arena.
\end{itemize}
For a detailed description of disturbances as well as hyperparameters for SMBO, see Section S3 of the Supplementary Information.

%% file: results.tex
\section{Results}
We now test the robot swarm on the foraging task. First, we analyse the behaviour-performance maps evolved by MAP-Elites. Second, we analyse adaptation, the potential to recover from adverse environmental perturbations.
\subsection{Evolution}
MAP-Elites has successfully evolved a large number of high-performing solutions (see Fig. S1 in Supplementary Information). For HBD, the global performance, the best performance across the entire behaviour-performance map, has converged to just above 5 food items per robot. For SDBC, the global performance is 4.5. A wide diversity of solutions is observable by the map coverage, with just above 100 solutions for SDBC and roughly 500 solutions for HBD. The lower average performance (just below 2 for HBD versus 1.3 for SDBC) illustrates that not all behaviours can be high-performing in the same environment. In sum, the HBD behaviour space yields high-quality maps and is chosen for the adaptation phase.

\begin{figure*}[htbp!]
\centering
\subfigure[Proximity sensor]{\includegraphics[width=0.22\textwidth]{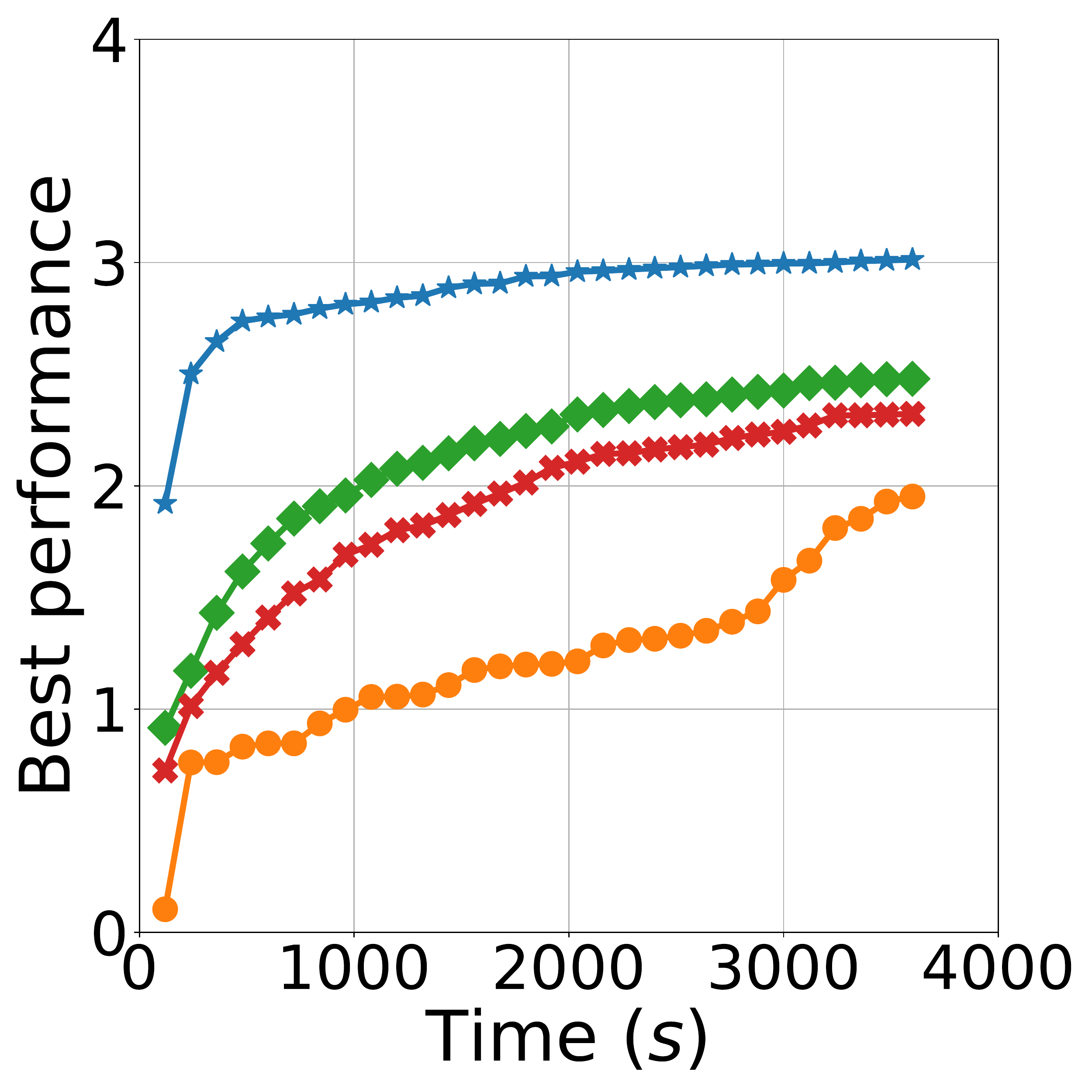}}  \hfill \subfigure[Actuator]{\includegraphics[width=0.22\textwidth]{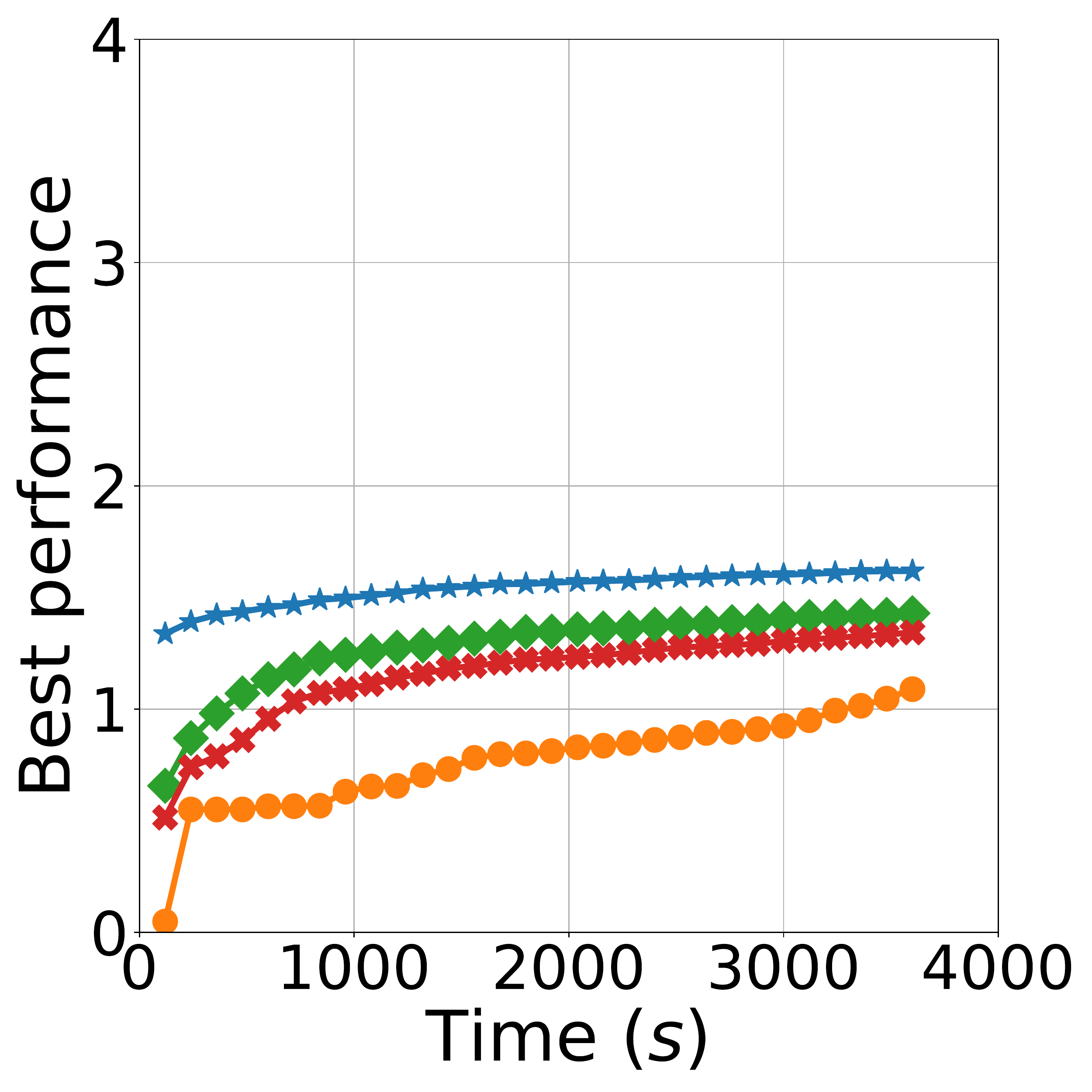}} \hfill
\subfigure[Software-food]{\includegraphics[width=0.22\textwidth]{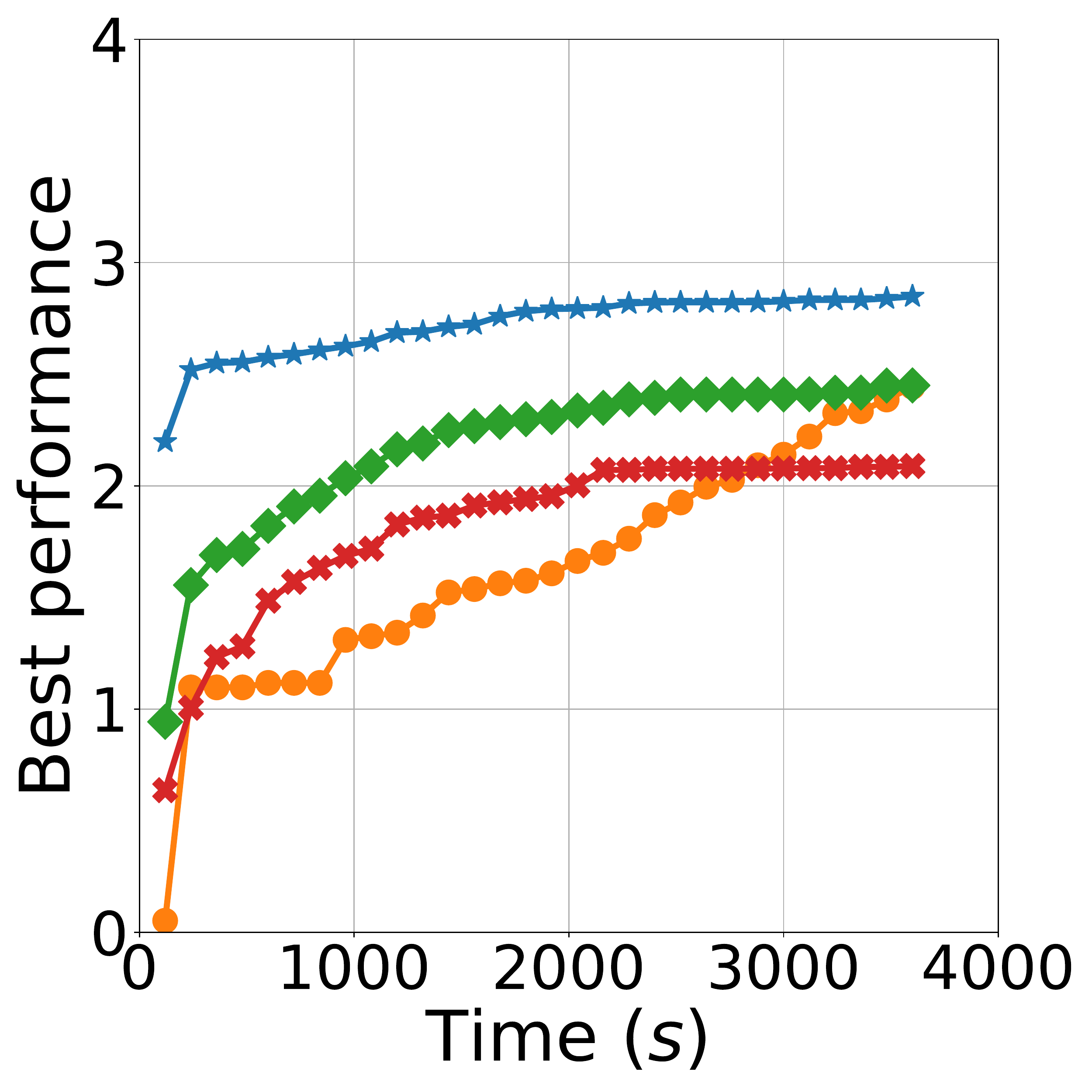}} \hfill
\subfigure[Food-scarcity]{\includegraphics[width=0.22\textwidth]{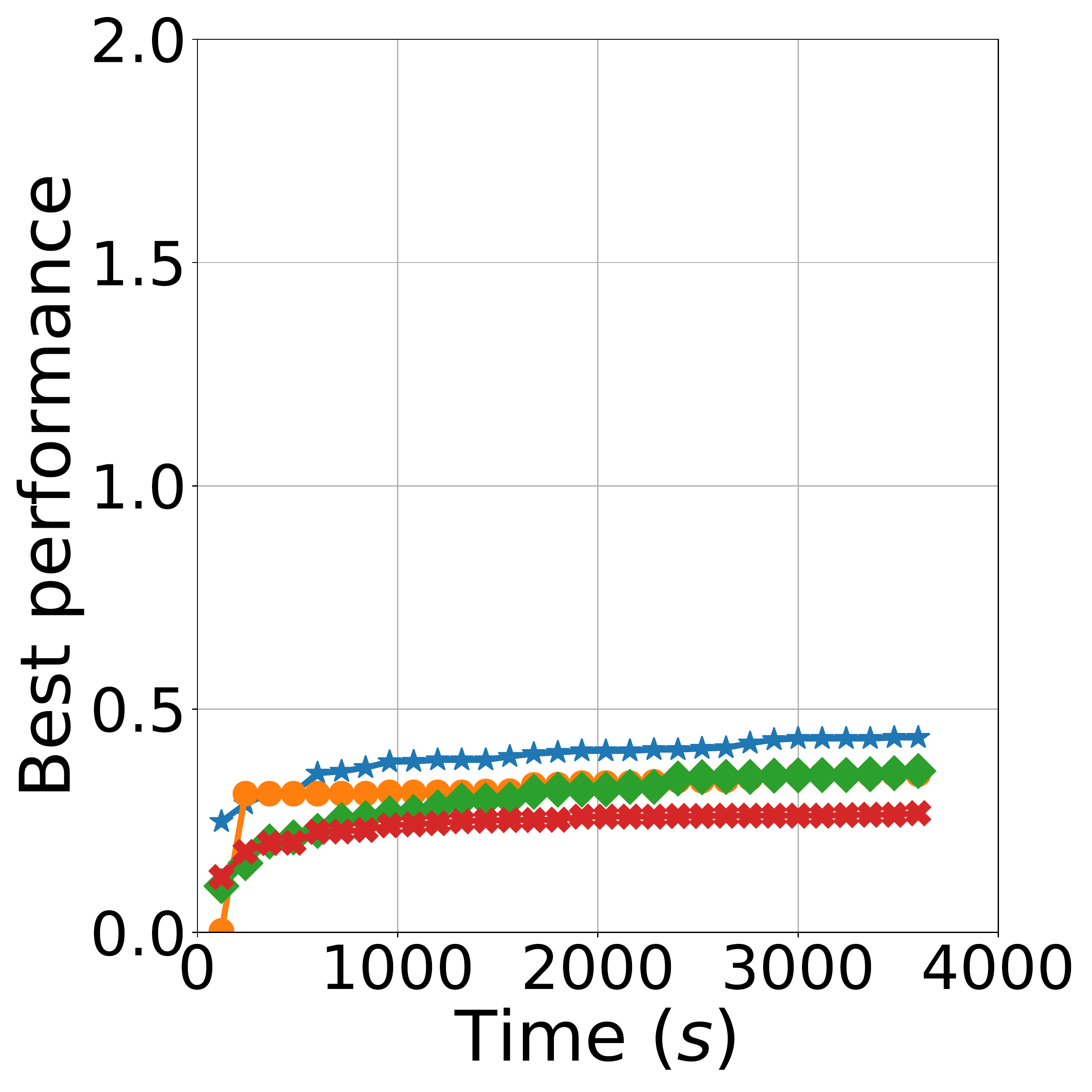}}\\
\includegraphics[scale=0.18]{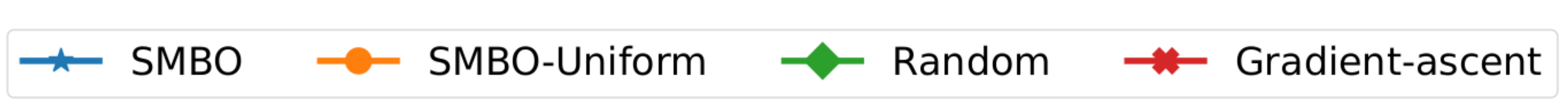}
\caption{\small{Development of performance (Mean aggregated across 5 replicates and all faults within the type of perturbation) as a function of the time consumption for centralised adaptation methods.} } \label{fig: development-homogeneous}
\end{figure*}
\subsection{Adaptation}
Using the HBD-based behaviour-performance maps, we now compare the performance of SMBO and the baseline behaviour search algorithms on recovery from perturbations to the robots or their surroundings. We distinguish between centralised adaptation, where each robot has a copy of the same controller, and decentralised adaptation, where robots are allowed to select their controller independently.\\
\indent \textbf{Centralised behaviour adaptation:} The SMBO learner is compared to random search and gradient ascent (see Section S2 of the Supplementary Information), and SMBO with a uniform prior set to the mean performance across the behaviour-performance map (\textbf{SMBO-Uniform}). Development of performance over performance evaluations (see Fig. \ref{fig: development-homogeneous}) illustrates that: (i) SMBO rapidly finds high-performing controllers for the homogeneous swarm, close to the maximal performance in the behaviour-performance map; (ii)  indicating the importance of a suitable prior for rapid adaptation, SMBO-Uniform initially cannot find a high-performing controller but eventually achieves a performance comparable to the first few evaluations of SMBO; (iii) the random search and gradient descent methods generally yield lower-performing controllers. Assessing the final performance obtained after 30 evaluations, the SMBO learner consistently outperforms the baseline algorithms across all 6 categories of disturbances (see Table \ref{tab: significance}). The smaller effect observed for the food scarcity condition may be because the prior from the behaviour-performance map is not as accurate; Fig. \ref{fig: development-homogeneous}d supports this explanation, because SMBO-Uniform improves equally quickly as SMBO.
\begin{table}[htbp!]
\centering
\caption{\small{Recovery performance (Mean $\pm$ SD) of SMBO compared to baseline algorithms after \SI{3600}{s}. Bold face indicates significance ($p<0.05$) with the Wilcoxon ranksum test, when compared to SMBO. Sign indicates the direction of the effect, (+) when SMBO is significantly higher and (-) when SMBO is significantly lower.}} \label{tab: significance}
\resizebox{0.48\textwidth}{!}{
\begin{tabular}{l  l l l l }
\toprule
\textbf{Disturbance}   & \multicolumn{4}{l}{\textbf{Learner}} \\
\midrule
&  \multicolumn{1}{c}{SMBO}& \multicolumn{1}{c}{SMBO-Uniform} & 
                     \multicolumn{1}{c}{Random} & \multicolumn{1}{c}{Gradient-ascent} \\
\hline
\textbf{Robot fault} &   &   &   &   \\ \hline
Proximity-Sensor & 
$3.01 \pm 1.24$   &$\mathbf{1.95 \pm 0.98}$ (+)  &$\mathbf{2.48 \pm 1.08}$ (+)  &$\mathbf{2.32 \pm 1.06}$ (+) \\
Ground-Sensor & 
$1.93 \pm 0.51$   &$1.84 \pm 0.38$  &$\mathbf{1.74 \pm 0.46}$ (+)  &$\mathbf{1.63 \pm 0.59}$ (+)  \\
Actuator & 
$1.62 \pm 0.75$   &$\mathbf{1.09 \pm 0.68}$ (+)  &$1.43 \pm 0.67$  &$\mathbf{1.34 \pm 0.78}$ (+)  \\
Software-Nest & 
$2.79 \pm 0.23$   &$\mathbf{2.30 \pm 0.41}$ (+)  &$\mathbf{2.39 \pm 0.23}$ (+)  &$\mathbf{2.16 \pm 0.56}$ (+) \\
Software-Food & 
$2.85 \pm 0.19$  &$\mathbf{2.44 \pm 0.49}$ (+)  &$\mathbf{2.45 \pm 0.27}$ (+)  &$\mathbf{2.09 \pm 0.68}$ (+)  \\
Food-Scarcity & 
$0.44 \pm 0.21$  &$0.35 \pm 0.17$  &$0.36 \pm 0.19$  &$\mathbf{0.27 \pm 0.20}$ (+) \\
\bottomrule
\end{tabular} 
}
\end{table}
\begin{table}
\centering
\caption{\small{Recovery performance (Mean $\pm$ SD) of SMBO-Decentralised compared to baseline algorithms after \SI{3600}{s}. In \textbf{(a)}, performance is based on the aggregate of best controllers during adaptation. In \textbf{(b)}, performance is calculated after joining together the best controllers found for each robot.}} \label{tab: significance-dec}
\subtable[Aggregate of best controllers during adaptation]{
\resizebox{0.48\textwidth}{!}{
\begin{tabular}{l  l l l l }
\toprule
\textbf{Disturbance}   & \multicolumn{4}{l}{\textbf{Learner}} \\
\midrule
&  \multicolumn{1}{c}{SMBO-Dec}& \multicolumn{1}{c}{SMBO-Dec Naive} & 
                     \multicolumn{1}{c}{SMBO No Sharing} & \multicolumn{1}{c}{Random No Sharing} \\
\hline
Proximity-Sensor & $3.30 \pm 1.51$  &$3.31 \pm 1.56$  &$\mathbf{2.91 \pm 1.51}$ (+)  &$\mathbf{1.20 \pm 0.90}$ (+)  \\
Food-Scarcity & $0.86 \pm 0.32$   &$0.88 \pm 0.35$  &$\mathbf{0.62 \pm 0.32}$ (+)  &$\mathbf{0.19 \pm 0.18}$ (+) \\
Proximity-Sensor 2x & 
$2.61 \pm 1.04$   &$2.59 \pm 1.06$  &$\mathbf{2.10 \pm 1.04}$ (+)  &$\mathbf{0.86 \pm 0.69}$ (+) \\
Food-Scarcity 2x & 
$0.86 \pm 0.29$  &$\mathbf{0.90 \pm 0.35}$ (-)  &$\mathbf{0.48 \pm 0.28}$ (+)  &$\mathbf{0.15 \pm 0.14}$ (+)  \\
\end{tabular}
}
}
\subtable[Composed swarm after adaptation]{
\resizebox{0.48\textwidth}{!}{
\begin{tabular}{l  l l l l }
\toprule
\textbf{Disturbance}   & \multicolumn{4}{l}{\textbf{Learner}} \\
\midrule
&  \multicolumn{1}{c}{SMBO-Dec}& \multicolumn{1}{c}{SMBO-Dec Naive} & 
                     \multicolumn{1}{c}{SMBO No Sharing} & \multicolumn{1}{c}{Random No Sharing} \\
\hline
Proximity-Sensor & $2.33 \pm 0.77$   &$2.30 \pm 0.75$  &$2.14 \pm 0.64$  &$\mathbf{1.76 \pm 0.60}$ (+)  \\
Food-Scarcity & $0.24 \pm 0.21$  &$0.26 \pm 0.24$  &$0.22 \pm 0.16$  &$0.21 \pm 0.15$ \\
Proximity-Sensor 2x & $1.45 \pm 0.32$  &$1.43 \pm 0.34$  &$\mathbf{1.36 \pm 0.26}$ (+)  &$\mathbf{1.16 \pm 0.26}$ (+) \\
Food-Scarcity 2x & 
$0.23 \pm 0.17$  &$0.20 \pm 0.15$  &$0.21 \pm 0.12$  &$0.19 \pm 0.12$ \\
\end{tabular}
}
}
\end{table}
\begin{figure*}[htbp!]
\centering
\subfigure[Proximity sensor]{\includegraphics[width=0.22\textwidth]{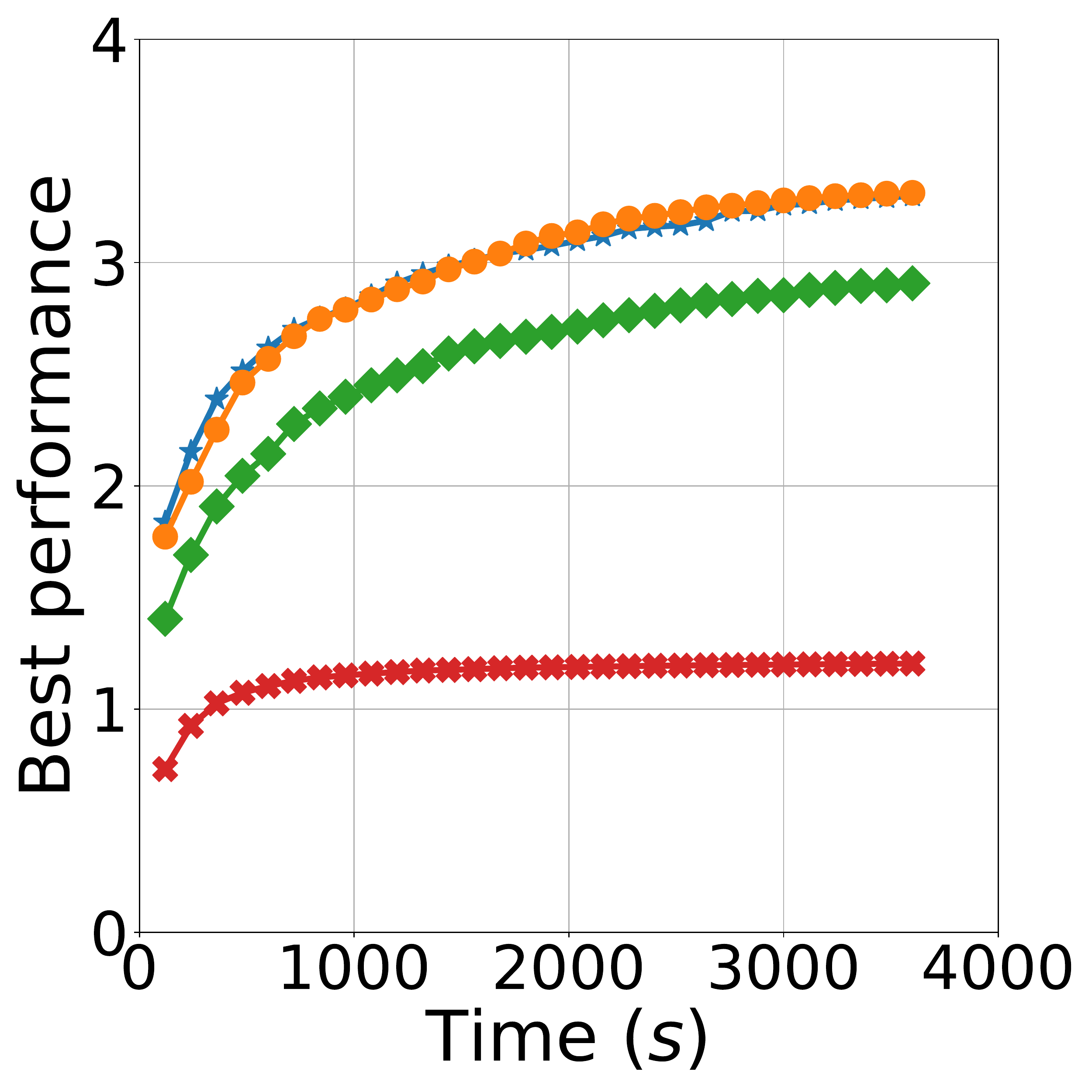}} \hfill \subfigure[Proximity sensor 2x]{\includegraphics[width=0.22\textwidth]{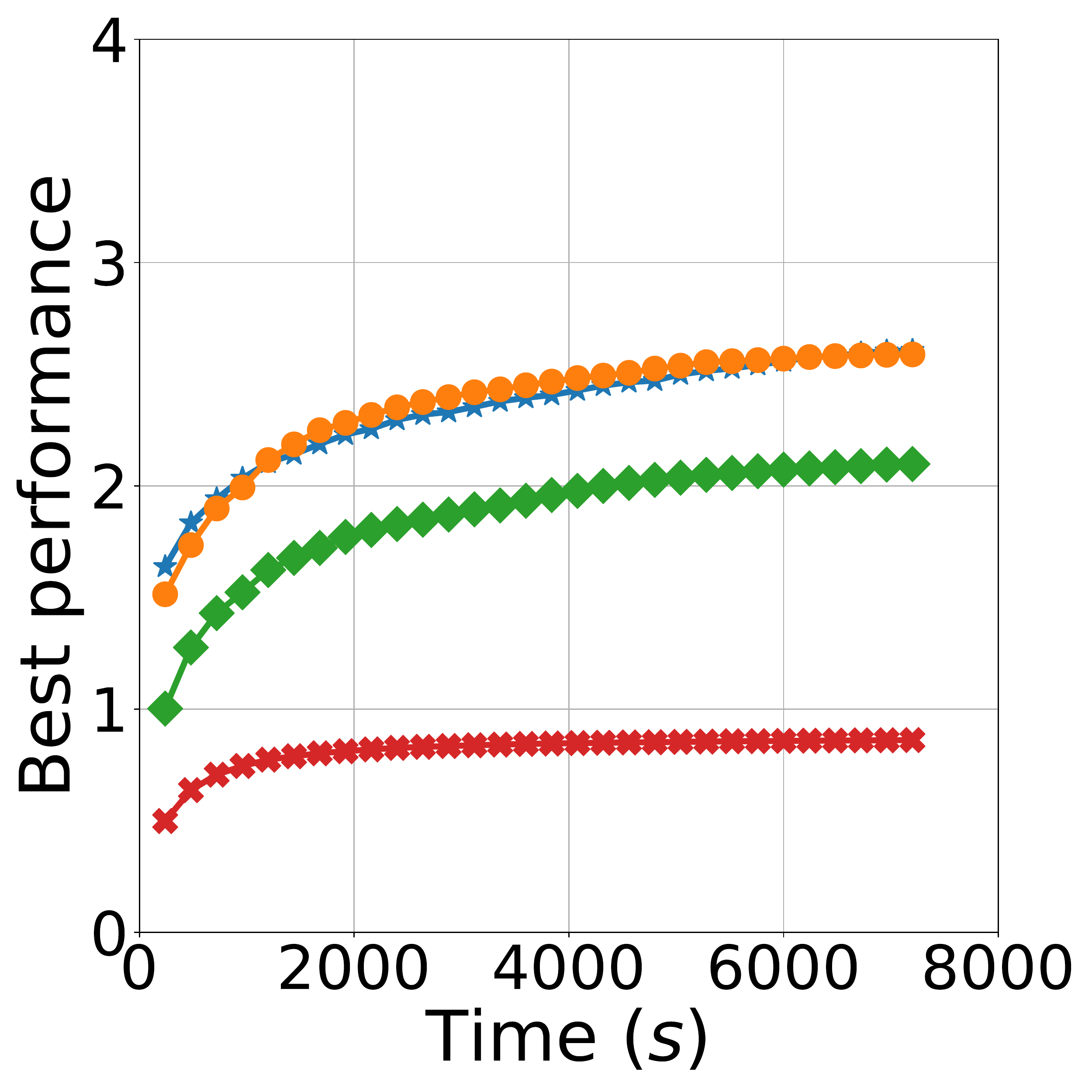}} \hfill \subfigure[Food-scarcity]{\includegraphics[width=0.22\textwidth]{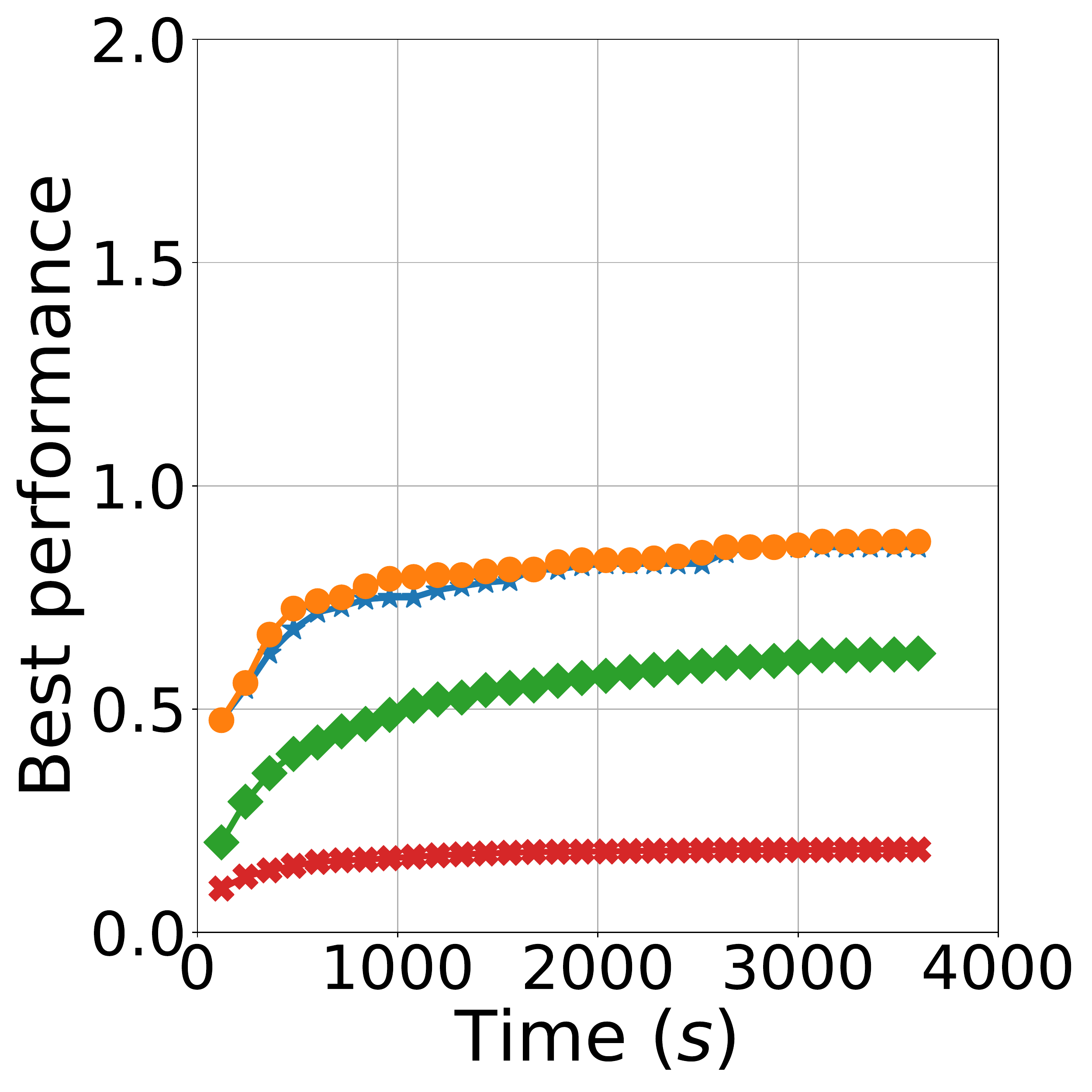}} \hfill \subfigure[Food-scarcity 2x]{\includegraphics[width=0.22\textwidth]{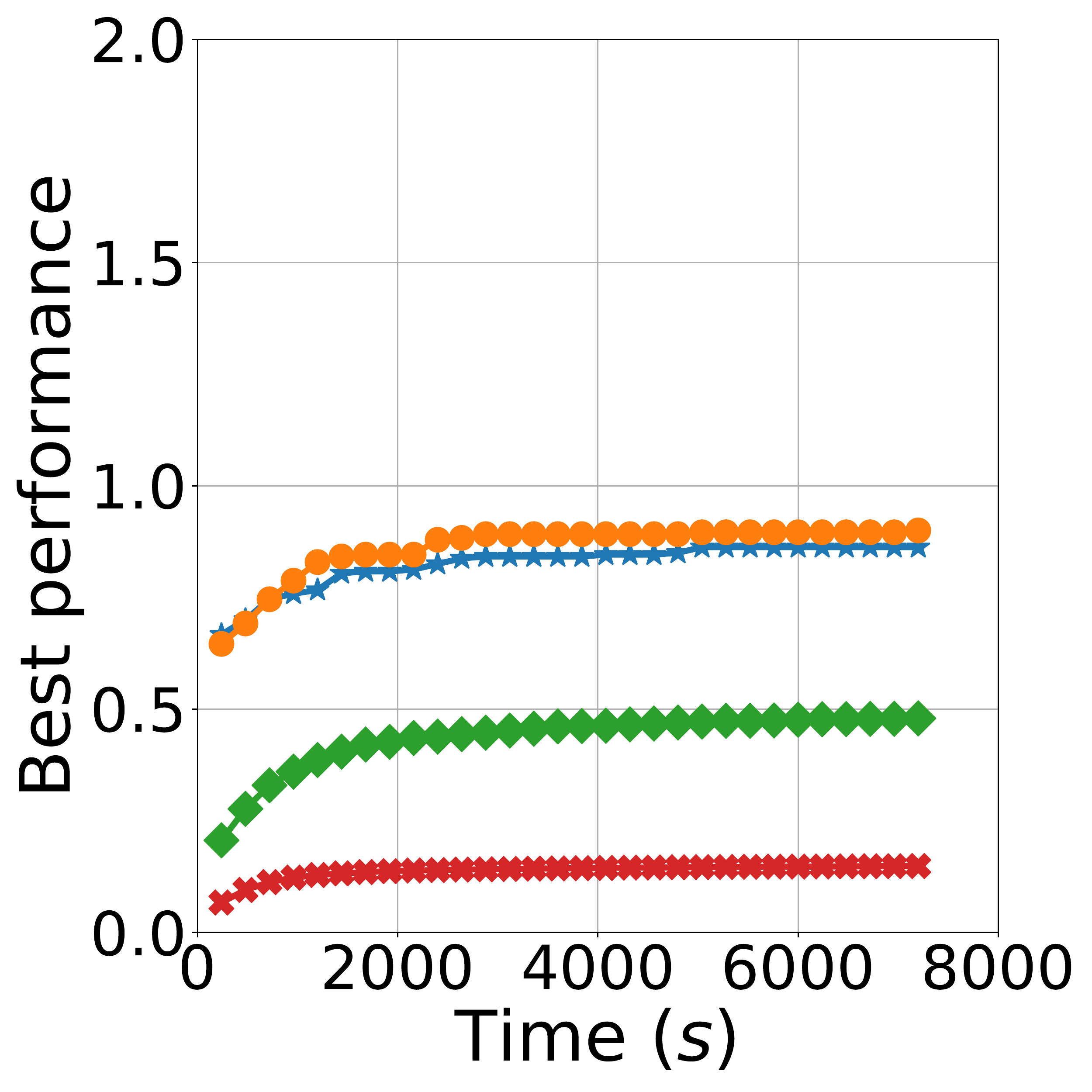}} \\
\includegraphics[scale=0.18]{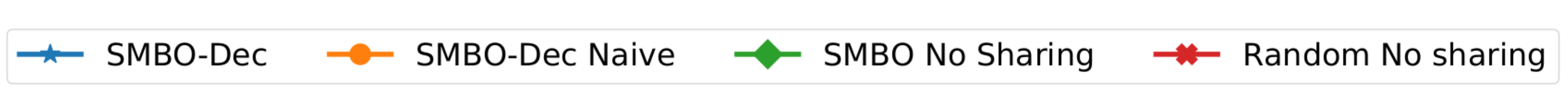}
\caption{\small{Development of performance (Mean aggregated across 5 replicates and all faults within the type of perturbation) as a function of the time consumption for decentralised adaptation methods.}} \label{fig: development-heterogeneous}
\end{figure*}

\indent \textbf{Decentralised behaviour adaptation:} In the decentralised experiments, SMBO-Decentralised (\textbf{SMBO-Dec}) is subjected to an ablation study where SMBO-Dec is gradually decomposed to study the impact of its different components. The impact of the local penalisation acquisition function is studied by comparison to a naive implementation of batch-based Bayesian optimisation, which we call \textbf{SMBO-Dec Naive}, in which the acquisition function is such that the first robot selects the controller with highest UCB, the second controller selects the controller with second-highest UCB, etc. To study the impact of sharing a Gaussian process model across robots, \textbf{SMBO No Sharing} further removes selective sharing of performance of evaluated controllers between robots with similar fault conditions. Finally, a random search, called \textbf{Random No Sharing}, further replaces the Bayesian optimisation component. To fully illustrate the advantages of the decentralised approach, we include experiments -- which we denote as \textbf{2x} -- where the swarm size is doubled as are the arena size and evaluation time per trial. We limit the number of experiments by focusing on one instance of fault recovery, proximity sensor faults, and one instance of surrounding change, food scarcity.\\
\indent Development of performance over time (see Fig. \ref{fig: development-heterogeneous}) illustrates that: (i) SMBO methods find controllers that are higher-performing than an exhaustive search across the behaviour-performance map for a homogeneous swarm; (ii) Random No Sharing performs similarly to the fault injection performance; (iii) SMBO-Dec methods perform higher than SMBO No Sharing; and (iv) there is a small benefit of SMBO-Dec over SMBO-Dec Naive but only in the initial phase of adaptation. While the larger arena of the 2x-tasks makes the problem more difficult\footnote{Although the evaluation time is scaled two times as well, the arena is different from the one in which the controllers were evolved and is larger compared to the robot body size, allowing more possible inefficient paths.}, resulting a lowered performance, comparison of the recovery performance to the fault injection performance illustrates now larger \% improvement. The statistical data of the final performance after \SI{3600}{s} (see Table \ref{tab: significance-dec}) is in line with our hypothesis: the performance decreases as more components of SMBO-Dec are ablated. During adaptation, SMBO-Dec yields a statistical significant effect compared to SMBO No Sharing and Random No Sharing (see Table \ref{tab: significance-dec}a). Forming a swarm from the best controllers reduces the performance but preserves the rank-order of the learners (see Table \ref{tab: significance-dec}b).

%% file: conclusion.tex
\section{Conclusion and future work}
In swarm robotics, rapid fault recovery and adaptation to unforeseen events remain a grand challenge. This paper demonstrates the use of Bayesian optimisation methods for rapid swarm behaviour adaptation. We propose: (i) a centralised algorithm called SMBO to adapt with a homogeneous centralised swarm; and (ii) a decentralised algorithm called SMBO-Decentralised to speed up the search of controllers by employing robots of the swarm as workers of a batch-based Bayesian optimisation algorithm. Application of SMBO in a real robot swarm could be achieved via synchronisation or consensus formation. In SMBO-Decentralised, each robot selectively shares a Gaussian process model with other comparable robots, such as those experiencing the same faults, and in this way, recovery can be greatly accelerated. Moreover, in SMBO-Decentralised each robot independently explores a different controller, providing a more robust communication framework since there are fewer failure modes.

Our foraging case study lays the foundations for rapidly adapting robot swarms. To expand the scope of our study, the following challenges need to be addressed:
\begin{itemize}
\item \textbf{Continuously changing environments}: the environment may be non-episodic, where consecutive evaluations are not identically and independently distributed and where no resets are performed. The asynchronicity of SMBO-Decentralised may be exploited to allow more frequent sharing of performance upon salient events that a robot may encounter. Continuous dynamics in the environment could be addressed by evolving dynamically changing policies such as plastic neural networks  \cite{Floreano2001,Risi2010b,Mouret2016} as the controllers in the behaviour-performance map.
\item \textbf{Representative robot performance estimates:} in SMBO-Decentralised, performance evaluations are based on a single robot that ignores the effects it has on the rest of the robot swarm. Some applications are not affected by this problem. For example, SMBO-Decentralised may recover single-robot gaits faster when multiple robots can gather samples at the same time without inter-robot interference. Another application is sparse robot swarms \cite{Tarapore2020}, the robots of which are spread across vast distances, such as is the case in coastal monitoring  or mapping a large forest. In dense robot swarms, the problem may be addressed by estimating the reliability of the performance evaluation by characterising the behaviour of neighbouring robots, and then using this information, e.g., by adding it to the standard-error in the kernel matrix. 
\end{itemize}